\documentclass[10pt,a4paper,twocolumn]{article}

\usepackage[T1]{fontenc}
\usepackage[utf8]{inputenc}
\usepackage{lmodern}
\usepackage[scaled=0.92]{inconsolata}
\usepackage{microtype}
\usepackage[margin=0.75in,columnsep=0.25in]{geometry}
\usepackage{amsmath,amssymb,amsfonts}
\usepackage{bm}
\usepackage[numbers,sort&compress]{natbib}
\usepackage{authblk}

\def\rmX{{\mathbf{X}}}

\def\mN{{\bm{N}}}

\DeclareMathAlphabet{\mathsfit}{\encodingdefault}{\sfdefault}{m}{sl}
\SetMathAlphabet{\mathsfit}{bold}{\encodingdefault}{\sfdefault}{bx}{n}

\def\gD{{\mathcal{D}}}

\def\sR{{\mathbb{R}}}

\usepackage{array}
\usepackage{booktabs}
\usepackage{multirow}
\usepackage{float}
\usepackage{xr}
\usepackage{xstring}
\usepackage[table]{xcolor}
\usepackage{xspace}
\usepackage{adjustbox}
\usepackage{graphicx}
\usepackage{enumerate}
\usepackage{enumitem}
\usepackage{caption}
\usepackage{subcaption}
\usepackage[colorlinks=true,allcolors=blue]{hyperref}
\usepackage[capitalise,noabbrev]{cleveref}
\newcommand{\mySqBullet}{\raisebox{0.25em}{{\scriptsize$_\blacksquare$}}}
\setlist[itemize]{label=\mySqBullet}

\newcommand{\ie}{i.e.\xspace}
\newcommand{\eg}{e.g.\xspace}

\newcommand{\polyview}{\texttt{polyview}\xspace}
\newcommand{\cmark}{\ensuremath{\checkmark}}
\newcommand{\xmark}{\ensuremath{\times}}
\newcommand{\stack}[1]{\begin{tabular}[c]{@{}c@{}}#1\end{tabular}}
\definecolor{supyes}{HTML}{C8E6C9}
\definecolor{suppart}{HTML}{FFE9A8}
\definecolor{supno}{HTML}{F8C9C4}
\newcommand{\yes}[1][\cmark]{\cellcolor{supyes}#1}
\newcommand{\partly}[1][partial]{\cellcolor{suppart}#1}
\newcommand{\no}[1][\xmark]{\cellcolor{supno}#1}
\newcommand{\legendbox}[2]{{\setlength{\fboxsep}{1pt}\colorbox{#1}{\strut #2}}}

\newcommand{\inlinetitle}[2]{%
  \par\vspace*{4pt}%
  \noindent\textbf{\emph{#1}{#2}}%
}

\newcommand{\NOTE}[1]{{\color{red} #1}}

\newcounter{marginNoteCounter}
\renewcommand{\mN}[1]		{\stepcounter{marginNoteCounter}\,{\footnotesize $^{\text{\color{red}\textbf\themarginNoteCounter}}$}\marginpar{\footnotesize $^\text{\themarginNoteCounter}$\,\NOTE{#1}}}
\newcounter{marginNoteCounterGw}

\title{polyview: A Python package for multi-view machine learning}
\author{Gwendal Debaussart-Joniec}
\author{Argyris Kalogeratos}
\affil{ENS Paris-Saclay, Gif-sur-Yvette, France}
\affil[ ]{\normalsize\texttt{\{gwendal.debaussart, argyris.kalogeratos\}@ens-paris-saclay.fr}}
\date{}

\begin{document}

\maketitle

\begin{abstract}
Multi-view learning jointly exploits multiple complementary representations of the same data and has become increasingly important in machine learning. However, the Python ecosystem lacks actively maintained, unified tooling for end-to-end multi-view workflows. In this paper, we present \polyview, a Python package that provides tools for multi-view embedding, clustering, fusion, and view augmentation, as well as for handling incomplete views, all compatible with scikit-learn. The library offers a unified interface for composing heterogeneous multi-view workflows, including seamless transitions between multi-view and single-view stages. It is built around a core set of classes and utilities that enable composition of different methods and straightforward implementation of new ones. We illustrate the package on five real multi-view datasets and compare its components based on canonical correlation analysis with those of two established libraries. 
\polyview aims to be both a practical toolkit for benchmarking and prototyping multi-view methods and a foundation for future research and development in this area.
\end{abstract}


\section{Introduction}

Advances in multimodal and distributed data acquisition 
have made multi-view datasets increasingly common. Each data object is described by several complementary representations, such as different modalities (\eg text, images, and sensor streams), acquisition instruments, or transformed versions of the same signal (\eg projections and feature decompositions). Multi-view learning aims to exploit these representations to improve robustness, interpretability, and downstream performance. Applications of multi-view learning span many domains, including social network analysis, multi-omics data, audio-visual data, and remote sensing \citep{bauvin2023multi,baiao2025survey,li2019survey,qin2025survey,xu2013multiviewsurvey,baltrusaitis2019multimodal,atrey2010multimodal, ghamisi2019multisource}.

Despite this growing demand, the Python ecosystem still lacks actively maintained, unified tooling for end-to-end multi-view workflows. The closest package in scope, \texttt{mvlearn}~\citep{perry2021mvlearn}, has had no update released since April 2022 and can no longer be installed on current Python versions, while \texttt{multiview}~\citep{multiview} has not been updated since 2018. Actively maintained alternatives, such as \texttt{cca-zoo}~\citep{ccazoo} or \texttt{multi-view-AE}~\citep{multiviewae}, focus on specific families of methods (embeddings based on canonical correlation analysis (CCA) and on autoencoders, respectively) and do not provide a unified interface for composing different multi-view strategies together, or with standard single-view preprocessing and learning methods. In addition, recent taxonomies of multi-view learning \citep{li2019survey} are not fully reflected in available libraries, where support for fusion methods and more-than-two-view settings remains limited.

For these reasons, we have developed \polyview, a Python package that brings the main multi-view strategies together behind the familiar scikit-learn (sklearn) API. It covers embedding, clustering, fusion, and view augmentation, and includes imputation tools for incomplete views. The package is designed to support practical composition in real workflows rather than isolated algorithm usage, with explicit support for bridging multi-view and single-view stages through \texttt{PolyPipeline}. \polyview prioritizes methods that are broadly applicable across heterogeneous data settings and arbitrary numbers of views. The open-source package is publicly available\footnote{\url{https://github.com/Gwendal-Debaussart/polyview}} under the GNU General Public License v3 (GPLv3), and the documentation is available online\footnote{\url{https://gwendal-debaussart.github.io/polyview/}}.

\section{Related software}\label{sec:related}

\begin{table*}[t]
  \centering
  \caption{Python packages for multi-view learning (September 2026). Colors: \legendbox{supyes}{supported}, \legendbox{suppart}{partly supported}, \legendbox{supno}{not supported}. \emph{MV/SV pipelines}: chaining multi-view and single-view stages. \emph{Works on}: \texttt{pip} install and import in a fresh Python~3.13\,/\,3.14 environment. \texttt{multiview} is marked partial as it declares no dependencies and these must be installed manually.
	}
  \label{tab:related}
  \footnotesize
  \setlength{\tabcolsep}{3pt}
  \setlength{\aboverulesep}{0pt}
  \setlength{\belowrulesep}{0pt}
  \renewcommand{\arraystretch}{1.3}
  \begin{tabular}{@{}lll|ccccccc}
    \toprule
    \textbf{Package} &\stack{\textbf{Latest} \\ \textbf{Release}} & \textbf{Main scope} & \stack{\textbf{$>$2}\\\textbf{views}} & \textbf{Clust.} & \textbf{Fusion} & \stack{\textbf{View}\\\textbf{augm.}} & \stack{\textbf{MV/SV}\\\textbf{pipelines}} & \stack{\textbf{sklearn}\\\textbf{API}} & \stack{\textbf{Works on}\\\textbf{3.13\,/\,3.14}} \\
    \midrule
    \texttt{mvlearn} & 0.5.0 (2022) & CCA variants, decomp. & \partly & \yes & \partly[early] & \yes & \partly & \yes & \no[\xmark\,/\,\xmark] \\
    \texttt{multiview} & 1.0 (2018) & MDS, t-SNE, spectral & \yes & \yes & \no & \no & \no & \partly & \partly[\cmark\,/\,\cmark] \\
    \texttt{cca-zoo} & 3.1.1 (2024) & CCA variants & \partly & \no & \no & \no & \no & \yes & \yes[\cmark\,/\,\cmark] \\
    \texttt{multi-view-AE} & 1.1.8 (2024) & Autoencoders & \partly & \no & \no & \no & \no & \partly & \no[\xmark\,/\,\xmark] \\
    \midrule
    \polyview & 0.1.0 (2026) & CCA, clustering, fusion & \yes & \yes & \yes[\stack{early\\kernel\\late}] & \yes & \yes & \yes & \yes[\cmark\,/\,\cmark] \\
    \bottomrule
  \end{tabular}
\end{table*}

\cref{tab:related} compares \polyview with the main Python packages for multi-view learning; each package was installed with \texttt{pip} and imported in a fresh environment on Python~3.13 and~3.14. Partial support denotes methods restricted to $2$ views, early fusion only, per-view wrappers only (MV/SV pipelines), sklearn naming without its \texttt{fit}/output signatures, or dependencies that must be installed manually.

\texttt{mvlearn}~\citep{perry2021mvlearn} is the closest package in scope, covering embedding, decomposition, clustering, and co-training, and its \texttt{compose} module provides per-view wrappers usable in sklearn pipelines. However, several of its methods are restricted to $2$ views, its last release dates from 2022, and its pin of \texttt{matplotlib}~($\leq$~3.3.4) makes it uninstallable on current Python versions. \texttt{multiview}~\citep{multiview} provides multi-view multidimensional scaling (MDS), t-distributed stochastic neighbor embedding (t-SNE), and spectral clustering, but has had a single release (2018) and declares no dependencies. \texttt{cca-zoo}~\citep{ccazoo} is actively maintained and sklearn-compatible, but restricted to CCA-family embeddings. \texttt{multi-view-AE}~\citep{multiviewae} implements multi-view autoencoders, but its pinned \texttt{pytorch-lightning} version fails at import time on current Python versions. \polyview is not intended to replace these specialized packages, but to complement them with a maintained, general-purpose framework in which heterogeneous multi-view strategies can be composed.

\section{The multi-view setting and strategy families}\label{sec:setting}

This section outlines the multi-view setting, its challenges, and the main algorithmic strategies used to address them. This framing motivates the design of \polyview and its task-oriented taxonomy. For a more detailed introduction to multi-view learning, we refer the reader to comprehensive surveys \citep{li2019survey,xu2013multiviewsurvey,baltrusaitis2019multimodal,atrey2010multimodal,qin2025survey,chen2022representation}.

Formally, the multi-view setting involves a dataset $\gD$ with $M$ different views (or modalities), $\gD= \{\rmX^{(1)}, \rmX^{(2)}, ..., \rmX^{(M)}\}$, where each $\rmX^{(m)} \in \sR^{N \times d_m}$ represents the same $N$ data objects in a potentially different feature space of dimension $d_m$. Where the data objects fall into groups, we write $K$ for their number, that is the number of classes or, in a clustering task, the number of clusters to be found. This means that, for an index $i=1,...,N$, the vectors $\rmX^{(1)}_i$, $\rmX^{(2)}_i$, ..., $\rmX^{(M)}_i$ correspond to the same underlying data object $i$ across different views. As mentioned, the views may be homogeneous or heterogeneous, possibly with radically different dimensionalities that make simple approaches (\eg view concatenation) problematic.
In some cases, the views may even be incomplete; in particular, some data objects may be missing in one or more views. \polyview handles this case through imputation or complete-case analysis (\cref{sec:limitations}). The goal of multi-view learning is to use the complementarity of the views to improve performance on downstream tasks such as clustering, classification, regression, or representation learning. This setting poses several challenges:
\begin{itemize}[leftmargin=1.5em, itemsep=0em, topsep=0.3em]
  \item \textbf{Shared latent structure.} Multiple views encode complementary information about a shared underlying phenomenon. The key challenge is formalizing and exploiting this shared structure while preserving view-specific signals.
  \item \textbf{Heterogeneous feature spaces.} Views may differ in dimensionality and modality (\eg continuous vs. discrete, dense vs. sparse), requiring learning mechanisms that operate across incompatible representation spaces.
  \item \textbf{Cross-view relationships.} Views exhibit correspondence or dependency structures (correlation, consistency, or conditional independence) that need to be discovered and leveraged.
  \item \textbf{Complementarity vs. redundancy trade-off.} Methods must balance exploiting overlapping (redundant) information and discovering unique (complementary) view contributions.
  \item \textbf{View quality and informativeness.} Although each view is often assumed to be sufficient on its own to provide learning results better than random guessing, views do not contribute equally when considered together; some may be noisy, redundant, or even misleading, which motivates adaptive weighting and selective integration.
\end{itemize}
The first point captures the core motivation for multi-view learning. However, this complementarity assumption does not always hold in practice, and view imbalance becomes the limiting factor.

Depending on the task and assumptions about the data (\ie native data or data augmentation), various multi-view approaches have been developed. 
In the rest of the paper,
we use the term ``workflow'' to refer to an end-to-end multi-view learning pipeline that may include multiple stages of preprocessing, view augmentation, representation learning, and a downstream task objective. These steps can be organized into four strategy families.

\inlinetitle{1. Representation alignment (MV $\rightarrow$ MV)}{.} Certain methods aim at finding one representation per view by using information from other views.\mN{this is the weaker developed case. add 1-2 sentences and 1-2 references. also in the previous version with the bullets (commented above), there was a disagreement in this first point between the bullet and the explanation paragraph (which I think says the correct point -- plz check)}

\inlinetitle{2. Fusion (MV $\rightarrow$ SV)}{.} Some methods merge views into a single shared representation that can then be used for downstream tasks. This family also includes \emph{late fusion} approaches that combine per-view predictions. Both are often grouped under \emph{multi-view representation learning} \citep{li2019survey}. The boundary between them is not always clear-cut in practice: for example, canonical correlation analysis (CCA) produces shared latent spaces (alignment), and among its multi-view extensions \citep{guo2021canonicalcorrelationanalysiscca}, some do focus on learning shared representations (\eg MCCA \citep{parra2018multisetcanonicalcorrelationanalysis}), while others focus on fusing views into a single ``well-aligned'' representation (\eg GCCA \citep{Tenenhaus_regularized_2011}).

\inlinetitle{3. Clustering (MV $\rightarrow$ labels)}{.} Some methods infer latent groups directly from multi-view data without necessarily computing an explicit shared representation \citep{chen2022representation,kumar2011coreg,gao2013multiviewnmf,cai2013multiview,kumar2011cotraining}.

\inlinetitle{4. View augmentation (SV $\rightarrow$ MV)}{.} Certain methods generate additional views from single-view data through projections, subspaces, or kernels, enabling multi-view analysis even when only one input view is provided. This is valuable for benchmarking on standard single-view datasets \citep{debaussartjoniec2025multiviewdiffusiongeometryusing,lindenbaum2020multiview} and for enhancing robustness when the original view is noisy \citep{papalexakis2013domoreviews,ho1998randomsubspace,bingham2001randomprojection,xu2013multiviewsurvey}.

In practice, these four strategy families are often combined within a single workflow and integrated with standard single-view preprocessing and learning methods (\cref{sec:application_usecase}). This taxonomy guides the design of \polyview by organizing implemented methods into submodules that reflect these strategies, while ensuring that they can be combined in a single pipeline.

\section{Package overview}

This section presents an overview of \polyview: the design principles that guided its development, the implemented components, and representative applications, with a focus on how the API supports composition into end-to-end workflows.

\subsection{Design principles}

\polyview is built to be a practical, extensible foundation for multi-view learning. To achieve this, the package is guided by four design principles:
\begin{itemize}[leftmargin=1.5em, itemsep=0em, topsep=0.3em]
  \item \textbf{Sklearn compatibility.} Estimators follow familiar sklearn conventions (hyperparameters set in the constructor, \texttt{fit} returning the estimator, \texttt{transform}/\texttt{predict}, fitted attributes with a trailing underscore) and inherit from sklearn's \texttt{BaseEstimator}, so they support \texttt{get\_params}, \texttt{set\_params}, and \texttt{clone}. Single-view sklearn estimators can be used directly inside \texttt{PolyPipeline}.
  \item \textbf{Flexibility.} Multi-view and single-view components can be combined in a single pipeline via \texttt{PolyPipeline}, enabling complex compositions without custom glue code.
  \item \textbf{Clarity.} Implementations prioritize readability and explicitness so that users can inspect, understand, and extend methods.
  \item \textbf{Applicability.} Included methods are selected to be broadly usable across heterogeneous data and, where relevant, across arbitrary numbers of views.
\end{itemize}
In practice, flexibility is ensured by clear base classes for multi-view methods, such as \texttt{BaseMultiViewClusterer} for clustering tasks, \texttt{BaseMultiViewEmbedder} for embedding tasks, or \texttt{BaseMultiViewTransformer} for general transformation tasks. This flexibility is further supported by the \texttt{MultiViewDataset} class, which provides a unified interface for handling multi-view data, as well as \texttt{PolyPipeline}, which allows users to compose multi-view and single-view components in a single workflow. Clarity is achieved through explicit implementations that closely follow the mathematical descriptions of the original methods, with comprehensive documentation linking code to methodological references.
The applicability criterion is especially important for practical adoption, as it avoids restricting the package to narrow, fixed-view use cases. Consequently, methods that are by design limited to $2$ views are not the primary focus.

\begin{figure*}[t]
  \centering
  \begin{subfigure}[t]{0.32\textwidth}
    \includegraphics[width=\textwidth]{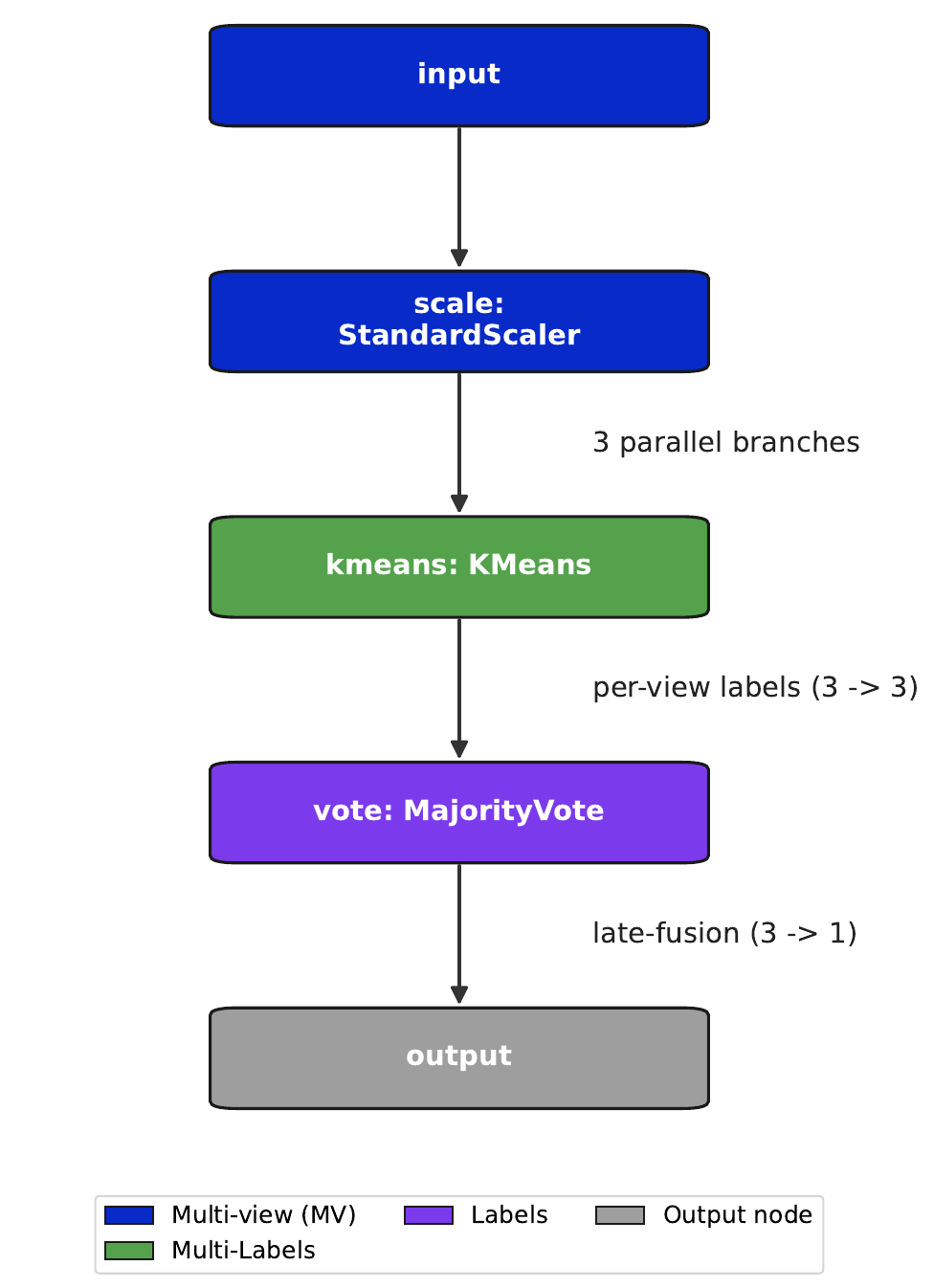}
    \caption{\textbf{Late fusion workflow:} parallel single-view steps followed by a multi-view fusion step.}
    \label{fig:majority_vote}
  \end{subfigure}
  \hfill
  \begin{subfigure}[t]{0.32\textwidth}
    \includegraphics[width=\textwidth]{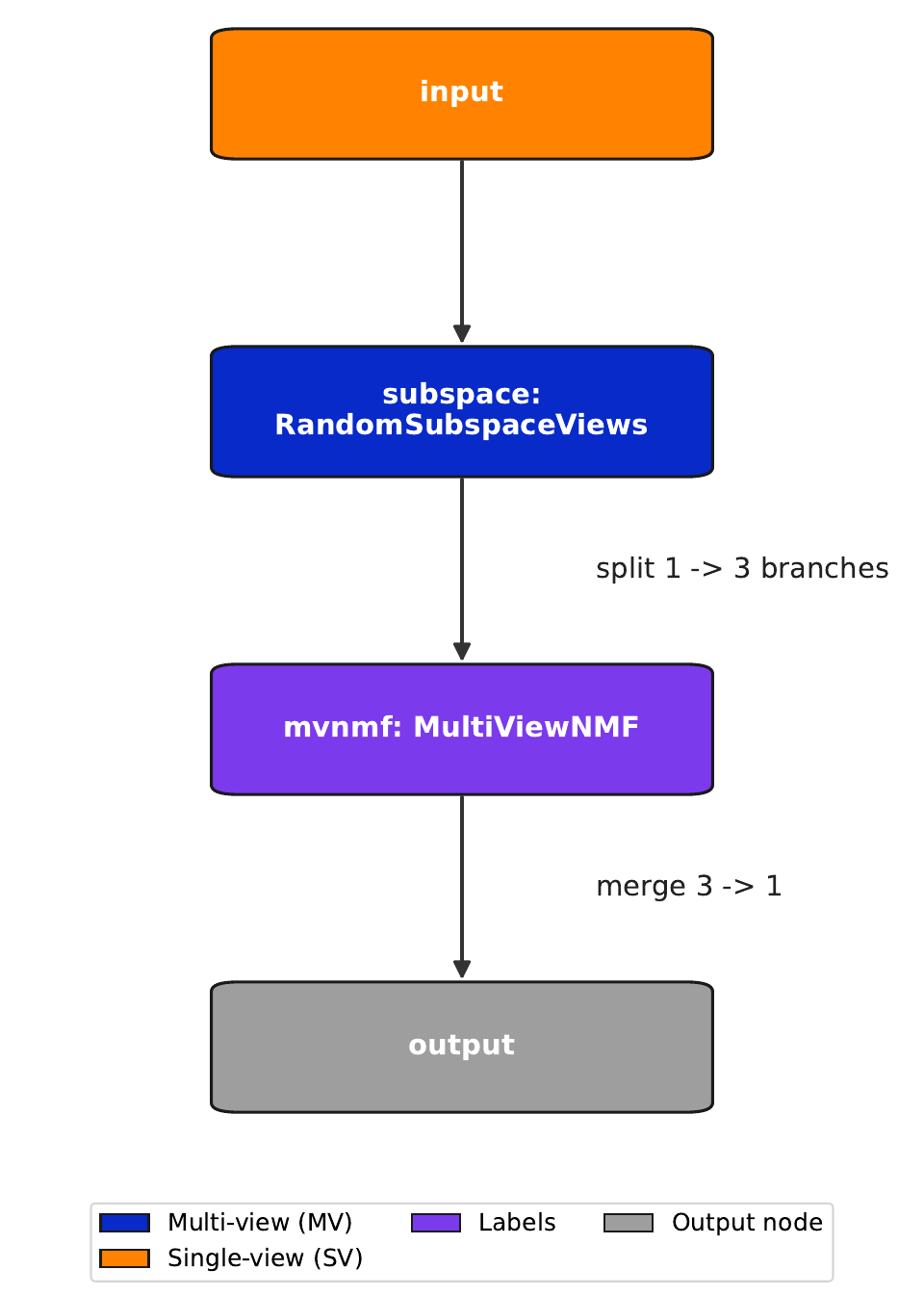}
    \caption{\textbf{Single-view to multi-view workflow:} single-view preprocessing followed by a multi-view clustering method.}
    \label{fig:sv_mv_clustering_pipeline}
  \end{subfigure}
  \hfill
  \begin{subfigure}[t]{0.32\textwidth}
    \includegraphics[width=\textwidth]{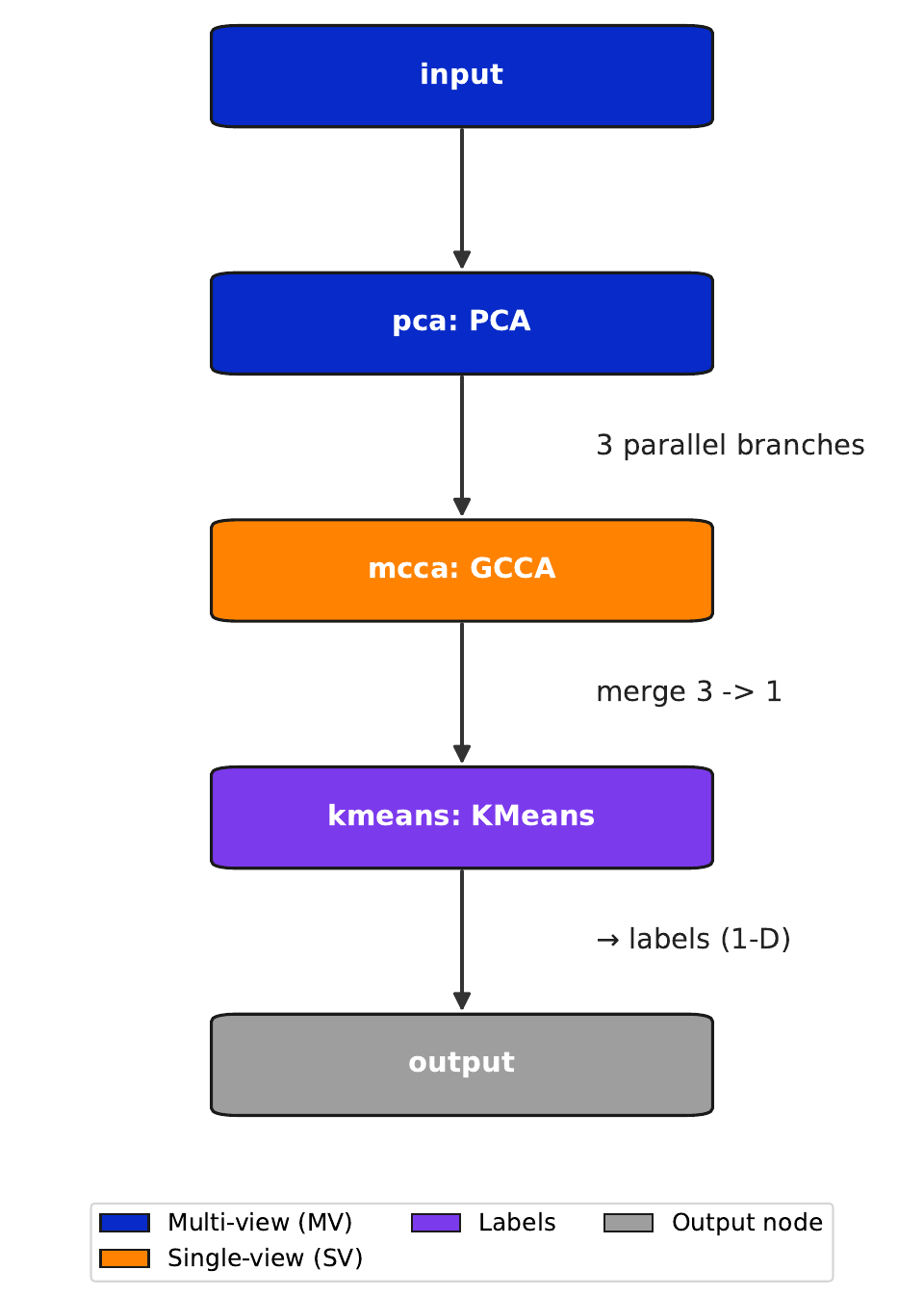}
    \caption{\textbf{Multi-view embedding workflow:} pre-processing steps followed by a multi-view embedding method and a clustering step.}
    \label{fig:mv_embedding_clustering_pipeline}
  \end{subfigure}
  \caption{Visualization of example \texttt{PolyPipeline} workflows, showing the sequence of transformations and the data flow through multi-view and single-view stages. The diagrams are generated using the \texttt{draw} method of \texttt{PolyPipeline}, which provides a visual representation of the pipeline structure. These examples are run on a $3$-view dataset or on a single-view dataset generating $3$ views through random subspaces.}
\end{figure*}

\subsection{Implemented components and limitations}\label{sec:limitations}

\begin{table}[t]
  \centering
  \caption{Method-level overview of implemented \polyview components, with representative references.}
  \label{tab:polyview-submodules}
  \begin{adjustbox}{width=\columnwidth,center}
  \begin{tabular}{lll}
    \toprule
    \textbf{Submodule} & \textbf{Method} & \textbf{Reference} \\
    \midrule
    \multirow{3}{*}{\texttt{embed}} & \texttt{GCCA} & \cite{carroll1968generalization,guo2021canonicalcorrelationanalysiscca} \\
    & \texttt{MCCA} & \cite{kettenring1971canonical,guo2021canonicalcorrelationanalysiscca} \\
    & \texttt{MultiViewMDS} & \cite{bai2017multidimensionalscalingmultipleinput} \\
    \midrule
    \multirow{4}{*}{\texttt{cluster}} & \texttt{MultiViewKMeans} & \cite{cai2013multiview,xu2013multiviewsurvey} \\
    & \texttt{MultiViewCoRegSpectralClustering} & \cite{kumar2011coreg,xu2013multiviewsurvey} \\
    & \texttt{MultiViewCoTrainSpectralClustering} & \cite{kumar2011cotraining,xu2013multiviewsurvey} \\
    & \texttt{MultiViewNMF} & \cite{gao2013multiviewnmf} \\
    \midrule
    \multirow{3}{*}{\texttt{augmentation}} & \texttt{MultiKernel} / \texttt{multi\_kernels} & \cite{gonen2011multiplekernel} \\
    & \texttt{RandomProjectionViews} / \texttt{random\_projection} & \cite{bingham2001randomprojection} \\
    & \texttt{RandomSubspaceViews} / \texttt{random\_subspace} & \cite{ho1998randomsubspace} \\
    \midrule
    \multirow{5}{*}{\texttt{fusion}} & \texttt{ConcatFusion} & \cite{atrey2010multimodal} \\
    & \texttt{WeightedFusion} & \cite{atrey2010multimodal} \\
    & \texttt{NormalizedFusion} & \cite{atrey2010multimodal} \\
    & \texttt{KernelFusion} & \cite{gonen2011multiplekernel} \\
    & \texttt{MajorityVote} & \cite{kuncheva2004combining} \\
    \midrule
    \multirow{4}{*}{\texttt{imputation}} & \texttt{SimpleViewImputer} & \cite{little2019statistical} \\
    & \texttt{KNNViewImputer} & \cite{troyanskaya2001missing} \\
    & \texttt{CrossViewRegressionImputer} & \cite{little2019statistical} \\
    & \texttt{DropIncompleteSamples} / \texttt{drop\_missing\_views} & \cite{little2019statistical} \\
    \bottomrule
  \end{tabular}
  \end{adjustbox}
\end{table}

As of the initial release, \polyview includes methods across the four strategy families outlined in \cref{sec:setting}. The package is organized into submodules that reflect these strategies while supporting straightforward composition within a single pipeline. In addition, the \texttt{imputation} submodule provides tools for handling incomplete views. These components are summarized in \cref{tab:polyview-submodules}.

\inlinetitle{Data model}{.} A multi-view dataset is passed either as a list of $2$-D arrays with the same number of rows, or as a \texttt{MultiViewDataset}, which also stores view names and labels and supports subsetting, (stratified) train/test splitting, and saving. Any number of views $M\geq 1$ of arbitrary dimensionalities is accepted.

\inlinetitle{Incomplete views}{.} Real multi-view datasets are often incomplete, in the particular sense that some data objects are not observed in every view. \polyview handles this setting through its \texttt{imputation} submodule, in which a missing view is encoded as a row of \texttt{NaN}. Missing views can be imputed i) from per-view statistics (\texttt{SimpleViewImputer}), ii) from the nearest samples that have the view, found using the views they share (\texttt{KNNViewImputer}), or iii) through a regression from the other views (\texttt{CrossViewRegressionImputer}); alternatively, they can be discarded through complete-case analysis (\texttt{DropIncompleteSamples}). Since imputers always return complete views, they can be placed as the first step of a \texttt{PolyPipeline}, after which any method of the package can be applied. Utilities to describe the missingness pattern and to simulate missing views, \eg for benchmarking, are also provided.

\inlinetitle{Model selection}{.} \texttt{PolyPipeline} exposes the parameters of its steps as \texttt{step\_\_parameter}, like sklearn's \texttt{Pipeline}. Passing the views as a single concatenated array, split back by a first \texttt{ViewSplitter} step, makes any \texttt{PolyPipeline} usable with \texttt{GridSearchCV} or \texttt{cross\_val\_score}.

\inlinetitle{Limitations}{.} The current release targets dense numerical features: sparse matrices and categorical variables must be handled by standard preprocessing beforehand. Incomplete views are only handled through imputation or complete-case analysis: none of the implemented methods learns directly from incomplete views, and such methods are planned for future releases.

\subsection{Example workflows} \label{sec:application_usecase}

Multi-view methods can be combined within a single workflow and integrated with standard single-view preprocessing and learning methods. Accordingly, the API of \polyview supports composition of different methods into end-to-end workflows. The library extends the principle of sklearn's \texttt{Pipeline} to multi-view settings through the \texttt{PolyPipeline} class, which allows users to chain multi-view and single-view components in one workflow. This design enables users to experiment with method combinations and iterate on workflows without writing custom glue code. Below, we provide representative use cases that illustrate the flexibility and applicability of \polyview's API. We use \texttt{PolyPipeline} to compose implemented methods into end-to-end workflows and \texttt{draw} to visualize the resulting pipelines.

\begin{table*}[t]
  \centering
  \caption{CCA-based embeddings of \polyview, \texttt{cca-zoo}, and \texttt{mvlearn}: median fit-and-transform time over $5$ runs, and NMI of $k$-means on the concatenated per-view projections (regularization $10^{-3}$; std.\ below $0.012$). The number of components is capped by the smallest view dimension ($6$, $7$, and $8$). $^\dagger$Python~3.9; no regularization for GCCA.}
  \label{tab:ccazoo}
  \footnotesize
  \setlength{\tabcolsep}{5pt}
  \begin{tabular}{llcccccc}
    \toprule
    & & \multicolumn{2}{c}{mfeat} & \multicolumn{2}{c}{Caltech101-7} & \multicolumn{2}{c}{Outdoor Scene} \\
    \cmidrule(lr){3-4}\cmidrule(lr){5-6}\cmidrule(lr){7-8}
    \textbf{Method} & \textbf{Library} & Time (s) & NMI & Time (s) & NMI & Time (s) & NMI \\
    \midrule
    \multirow{3}{*}{GCCA} & \polyview & 0.04 & 0.768 & 0.38 & 0.481 & 0.23 & 0.510 \\
    & \texttt{cca-zoo} & 0.53 & 0.768 & 1.94 & 0.480 & 1.47 & 0.507 \\
    & \texttt{mvlearn}$^\dagger$ & 0.52 & 0.800 & 3.20 & 0.473 & 0.96 & 0.444 \\
    \midrule
    \multirow{3}{*}{MCCA} & \polyview & 0.02 & 0.769 & 1.55 & 0.477 & 0.09 & 0.486 \\
    & \texttt{cca-zoo} & 0.13 & 0.767 & 6.18 & 0.479 & 0.48 & 0.491 \\
    & \texttt{mvlearn}$^\dagger$ & 0.38 & 0.770 & 2.76 & 0.481 & 0.64 & 0.484 \\
    \bottomrule
  \end{tabular}
\end{table*}

\begin{table}[t]
  \centering
  \caption{Fit time (in seconds, single run) on synthetic data with $M=3$ views of $50$ features and $5$ clusters, for an increasing number of samples $N$.}
  \label{tab:scaling}
  \footnotesize
  \setlength{\tabcolsep}{3pt}
  \begin{tabular}{@{}lrrrr@{}}
    \toprule
    \textbf{Method} & $N=500$ & $1000$ & $2000$ & $4000$ \\
    \midrule
    \texttt{ConcatFusion} + \texttt{KMeans} & 0.011 & 0.019 & 0.036 & 0.062 \\
    \texttt{MultiViewKMeans} & 0.17 & 0.24 & 0.64 & 1.12 \\
    \texttt{MultiViewNMF} & 0.48 & 0.87 & 1.57 & 2.93 \\
    \texttt{MCCA} & 0.002 & 0.003 & 0.004 & 0.004 \\
    \texttt{GCCA} & 0.016 & 0.023 & 0.023 & 0.037 \\
    \texttt{MultiViewCoRegSpectralClustering} & 0.053 & 0.17 & 0.83 & 4.10 \\
    \texttt{MultiViewMDS} & 0.90 & 3.72 & 15.8 & 65.1 \\
    \bottomrule
  \end{tabular}
\end{table}

\cref{fig:majority_vote} illustrates a simple workflow that applies multiple single-view steps in parallel to each view (\texttt{StandardScaler} and \texttt{KMeans}), and then combines their predictions through a multi-view fusion step (\texttt{MajorityVote}). Since cluster identifiers are arbitrary in each view, \texttt{align\_labels=True} first matches the per-view labels (Hungarian algorithm) before voting. \cref{fig:mv_embedding_clustering_pipeline} illustrates a workflow that first applies a preprocessing PCA step to each view, then applies a multi-view embedding method (GCCA) that learns a shared latent space across views, and finally applies a single-view clustering method (KMeans) to the resulting shared representation. Additional detailed runnable examples are provided in the package documentation and are intended to serve both as quick-start recipes and as templates for custom pipelines.

\section{Empirical validation}\label{sec:experiments}

We validate \polyview against established libraries on real multi-view data, report runtimes, and compare representative workflows built with the package.

\inlinetitle{Setup}{.} We use $3$ standard benchmarks: UCI Multiple Features (mfeat)~\citep{duin1998mfeat}, with $2000$ handwritten digits, $10$ classes, and $6$ views of $6$ to $240$ features; Caltech101-7~\citep{feifei2004caltech,li2015bipartite}, with $1474$ images, $7$ classes, and $6$ views of $40$ to $1984$ features; and Outdoor Scene~\citep{oliva2001scene}, with $2688$ images, $8$ classes, and $4$ views of $48$ to $512$ features. Views are standardized. Each embedding is evaluated by the NMI of $k$-means (with $K$ the number of classes) on the concatenated per-view projections, averaged over $5$ seeds, and by its median fit-and-transform time over $5$ runs. The experiments were run on an Apple M3 Max laptop with Python~3.13; \texttt{mvlearn}, which cannot be installed on this version, was run with Python~3.9 and the library versions current at its last release.

\inlinetitle{Comparison with \texttt{cca-zoo} and \texttt{mvlearn}}{.} We compare GCCA and MCCA with their counterparts using the same number of components, regularization, and downstream $k$-means (\cref{tab:ccazoo}). \polyview and \texttt{cca-zoo} reach the same NMI (differences $\leq 0.005$), as do the three MCCA implementations ($\leq 0.007$), which supports the correctness of the implementations. 
The GCCA of \texttt{mvlearn} relies on an unregularized formulation, which helps on mfeat ($0.800$ against $0.768$) but hurts on the other two datasets ($0.473$ against $0.481$, and $0.444$ against $0.510$). \polyview is the fastest in all settings, by a factor of $4$ to $13$ against \texttt{cca-zoo} and $2$ to $19$ against \texttt{mvlearn} (whose timings, obtained with older software, should be compared with caution). The smallest gain is MCCA on Caltech101-7, the dataset with by far the widest views, where \polyview takes $1.55$\,s against $2.76$\,s for \texttt{mvlearn}.

\inlinetitle{Scalability}{.} \cref{tab:scaling} reports fit times on synthetic data (\texttt{make\_multiview\_gaussian}). Early fusion, \texttt{MultiViewKMeans}, and \texttt{MultiViewNMF} scale roughly linearly with $N$, and GCCA and MCCA stay below $0.05$\,s up to $N=4000$. \texttt{MultiViewCoRegSpectralClustering} and \texttt{MultiViewMDS} operate on dense $N\times N$ matrices, so their time and memory grow at least quadratically; they are best suited to datasets of up to a few thousand samples.

\begin{table*}[t]
  \centering
  \newcommand{\std}[1]{{\tiny$\,\pm$#1}}
  \caption{Clustering results of \polyview workflows: AMI (mean $\pm$ standard deviation over $5$ seeds) and median wall-clock time (s) of the complete workflow. Best AMI and best time per dataset in bold. \emph{Best single view}: selected a posteriori using the labels. \emph{View augmentation}: $3$ views generated from the concatenated views (random subspaces of half of the features or random projections to $50$ dimensions).}
  \label{tab:workflows}
  \scriptsize
  \setlength{\tabcolsep}{2.5pt}
  \begin{adjustbox}{max width=\textwidth}
  \begin{tabular}{@{}lcccccccccc@{}}
    \toprule
    & \multicolumn{2}{c}{\stack{mfeat\\($N{=}2000$, $M{=}6$, $K{=}10$)}} & \multicolumn{2}{c}{\stack{Caltech101-7\\($N{=}1474$, $M{=}6$, $K{=}7$)}} & \multicolumn{2}{c}{\stack{Outdoor Scene\\($N{=}2688$, $M{=}4$, $K{=}8$)}} & \multicolumn{2}{c}{\stack{MSRC-v5\\($N{=}210$, $M{=}5$, $K{=}7$)}} & \multicolumn{2}{c}{\stack{100leaves\\($N{=}1600$, $M{=}3$, $K{=}100$)}} \\
    \cmidrule(lr){2-3}\cmidrule(lr){4-5}\cmidrule(lr){6-7}\cmidrule(lr){8-9}\cmidrule(lr){10-11}
    \textbf{Workflow} & AMI & Time & AMI & Time & AMI & Time & AMI & Time & AMI & Time \\
    \midrule
    \multicolumn{11}{l}{\emph{Single view and fusion}} \\
    Best single view + \texttt{KMeans} & 0.748\std{0.025} & 0.37 & 0.542\std{0.003} & 0.53 & 0.451\std{0.013} & 0.56 & 0.562\std{0.019} & 0.05 & 0.690\std{0.005} & 0.54 \\
    \texttt{ConcatFusion} + \texttt{KMeans} & 0.812\std{0.022} & 0.13 & \textbf{0.562}\std{0.004} & 0.42 & 0.565\std{0.003} & 0.43 & 0.675\std{0.018} & \textbf{0.03} & 0.916\std{0.011} & 0.20 \\
    Per-view \texttt{KMeans} + \texttt{MajorityVote} & 0.440\std{0.048} & 0.42 & 0.249\std{0.043} & 0.59 & 0.218\std{0.030} & 0.67 & 0.296\std{0.035} & 0.07 & 0.464\std{0.006} & 0.53 \\
    \quad with \texttt{align\_labels=True} & 0.747\std{0.037} & 0.42 & 0.439\std{0.016} & 0.58 & 0.383\std{0.012} & 0.62 & 0.408\std{0.028} & 0.06 & 0.619\std{0.012} & 0.54 \\
    \midrule
    \multicolumn{11}{l}{\emph{View augmentation + \texttt{MultiViewKMeans}}} \\
    \texttt{RandomSubspaceViews} & 0.829\std{0.023} & 6.96 & 0.551\std{0.050} & 24.4 & \textbf{0.566}\std{0.005} & 27.4 & 0.656\std{0.006} & 0.70 & 0.859\std{0.007} & 22.0 \\
    \texttt{RandomProjectionViews} & 0.794\std{0.030} & 1.34 & 0.533\std{0.031} & 0.92 & 0.532\std{0.013} & 2.23 & 0.638\std{0.024} & 0.11 & 0.860\std{0.007} & 15.4 \\
    \midrule
    \multicolumn{11}{l}{\emph{Multi-view clustering}} \\
    \texttt{MultiViewKMeans} & 0.680\std{0.005} & 8.47 & 0.379\std{0.025} & 23.2 & 0.479\std{0.005} & 18.1 & 0.551\std{0.047} & 0.62 & 0.795\std{0.013} & 16.0 \\
    \texttt{MultiViewCoRegSpectralClustering} & \textbf{0.900}\std{0.001} & 2.72 & 0.511\std{0.002} & 1.31 & 0.545\std{0.001} & 3.59 & \textbf{0.715}\std{0.013} & 0.06 & 0.890\std{0.005} & 27.9 \\
    \texttt{MultiViewCoTrainSpectralClustering} & 0.842\std{0.000} & 4.09 & 0.451\std{0.006} & 1.84 & 0.407\std{0.030} & 2.21 & 0.671\std{0.005} & 0.16 & \textbf{0.930}\std{0.011} & 7.21 \\
    \texttt{MultiViewNMF} & 0.503\std{0.030} & 6.35 & 0.469\std{0.031} & 30.7 & 0.362\std{0.017} & 19.9 & 0.525\std{0.031} & 1.57 & 0.477\std{0.019} & 5.30 \\
    \midrule
    \multicolumn{11}{l}{\emph{Multi-view embedding + \texttt{KMeans}}} \\
    \texttt{GCCA} & 0.761\std{0.001} & 0.08 & 0.472\std{0.001} & \textbf{0.41} & 0.469\std{0.003} & 0.26 & 0.355\std{0.030} & \textbf{0.03} & 0.703\std{0.008} & \textbf{0.19} \\
    \texttt{MCCA} & 0.767\std{0.001} & \textbf{0.06} & 0.473\std{0.001} & 1.55 & 0.483\std{0.012} & \textbf{0.16} & 0.378\std{0.012} & 0.22 & 0.759\std{0.012} & 0.23 \\
    \texttt{MultiViewMDS} & 0.592\std{0.014} & 20.9 & 0.336\std{0.008} & 11.5 & 0.349\std{0.001} & 33.5 & 0.423\std{0.028} & 0.30 & 0.576\std{0.005} & 10.7 \\
    \bottomrule
  \end{tabular}
  \end{adjustbox}
\end{table*}

\inlinetitle{Workflow comparison}{.} \cref{tab:workflows} reports clustering results for representative \polyview workflows, on the three datasets above and on two further benchmarks: MSRC-v5~\citep{winn2005msrc}, with $210$ images, $7$ classes, and $5$ views of $24$ to $576$ features, and 100leaves~\citep{mallah2013leaves}, with $1600$ plant leaves, $100$ classes, and $3$ views of $64$ features (shape, margin, and texture). Views are standardized (min--max scaled for \texttt{MultiViewNMF}, which requires non-negative inputs), and the number of clusters is set to the number of classes ($k=K$). Every workflow is expressed with the same API, either as a single multi-view estimator or as a \texttt{PolyPipeline} combining \polyview and sklearn components.

Which is the most effective strategy in each case depends on the dataset: looking at the AMI index, \texttt{MultiViewCoRegSpectralClustering} is best on mfeat ($0.90$) and MSRC-v5 ($0.72$), \texttt{MultiViewCoTrainSpectralClustering} on 100leaves ($0.93$), and early fusion and random-subspace views on Caltech101-7 and Outdoor Scene. This echoes the view-imbalance issue of \cref{sec:setting} and motivates comparing strategies within a single interface. Late fusion also illustrates the importance of label matching, as \texttt{align\_labels=True} raises the AMI of majority voting from $0.22$--$0.46$ to $0.38$--$0.75$. The fastest workflows are consistently early fusion and the CCA-based embeddings.

\section{Software quality, availability, and extensibility}\label{sec:software}

\inlinetitle{Installation and testing}{.} \polyview is a pure-Python package installed with \texttt{pip install polyview}. The package's dependencies are NumPy, SciPy, and scikit-learn (matplotlib and NetworkX are optional, for visualizations with \texttt{draw}). The version described here is 0.1.0. Its $361$ unit tests cover every estimator, \texttt{MultiViewDataset}, \texttt{PolyPipeline}, and the imputation tools, for $94\%$ line coverage. The suite passes on Python~3.13 and~3.14. All stochastic components expose a \texttt{random\_state} parameter, allowing for pipeline control.

\inlinetitle{Documentation}{.} The Sphinx documentation includes an API reference citing the original publication of each method and worked examples of each workflow type (MV$\to$MV, MV$\to$SV, SV$\to$MV, and late fusion). Contribution guidelines describe the coding conventions and the API contract for new estimators, and GitHub Actions build the documentation and publish releases to PyPI.

\inlinetitle{Extending \polyview}{.} A new method subclasses the base class of its task, which provides input validation and parameter handling, and tells \texttt{PolyPipeline} how the step changes the data. For instance, a minimal early-fusion method reads:
\vspace{0.5em}
{\small
\hrule
\begin{verbatim}
class MyConcatFusion(BaseFusion):
    def fit(self, views, y=None):
        self._validate_views(views)
        return self

    def transform(self, views):
        views = self._validate_views(
            views, reset=False)
        return np.hstack(views)
\end{verbatim}
\hrule
}

\section{Conclusion and future work}\label{sec:conclusion}

In this paper, we introduced \polyview, which provides a unified multi-view learning framework that covers embedding, clustering, fusion, view augmentation, and incomplete views. Its \texttt{PolyPipeline} class explicitly bridges multi-view and single-view stages, so that multi-view methods can be combined with standard sklearn components. Our experiments on three real multi-view datasets show that the CCA-based components of \polyview reach a clustering quality comparable to that of \texttt{cca-zoo} and \texttt{mvlearn}, with lower runtimes.

Future work will add methods that learn directly from incomplete views, improve the scalability of methods relying on dense $N\times N$ matrices, expand the set of implemented methods and the documentation, and encourage contributions from the community.

\section*{Acknowledgments}
This work was supported by the Industrial Data Analytics and Machine Learning Chair hosted at ENS Paris-Saclay.

{\small
\setlength{\bibsep}{0pt plus 0.3ex}
\bibliographystyle{abbrvnat}
\bibliography{bibliography}
}

\end{document}